\documentclass[letterpaper, 10 pt, conference]{ieeeconf}  

\IEEEoverridecommandlockouts                              

\usepackage{times}
\usepackage{epsfig}
\usepackage{graphicx}
\usepackage{amsmath}
\usepackage{amssymb}
\usepackage{color}
\usepackage{multirow}
\usepackage{tabulary}
\usepackage{tabularx}
\usepackage{pifont}
\usepackage{cite}
\usepackage{adjustbox}
\usepackage{xspace}
\usepackage{mathtools}

\makeatletter
\DeclareRobustCommand\onedot{\futurelet\@let@token\@onedot}
\def\@onedot{\ifx\@let@token.\else.\null\fi\xspace}
\def\eg{\emph{e.g}\onedot} 
\def\ie{\emph{i.e}\onedot}

\makeatother

\definecolor{mygray}{gray}{0.75}
\newcommand{\bd}[1]{\textbf{#1}}
\newlength\savewidth\newcommand\shline{\noalign{\global\savewidth\arrayrulewidth
  \global\arrayrulewidth 1pt}\hline\noalign{\global\arrayrulewidth\savewidth}}
\newcommand{\cmark}{\ding{51}}
\newcommand{\xmark}{\ding{55}}

\title{\LARGE \bf
End-to-end Contextual Perception and Prediction\\with Interaction Transformer}

\author{Lingyun (Luke) Li$^{1}$ \quad Bin Yang$^{1,2}$ \quad Ming Liang$^{1}$ \quad Wenyuan Zeng$^{1,2}$ \quad Mengye Ren$^{1,2}$\\
Sean Segal$^{1,2}$ \quad Raquel Urtasun$^{1,2}$ \\
\thanks{$^{1}$Uber Advanced Technologies Group}%
\thanks{$^{2}$University of Toronto}%
}

\begin{document}
\maketitle
\begin{abstract}
In this paper, we tackle the problem of detecting objects in 3D and forecasting their future motion in the context of self-driving. 
Towards this goal, we design a novel approach that explicitly takes into account the interactions between actors.
To capture their spatial-temporal dependencies, we propose a recurrent neural network with a novel Transformer \cite{transformer} architecture, which we call the \emph{Interaction Transformer}.
Importantly, our model can be trained end-to-end, and runs in real-time. 
We validate our approach on two challenging real-world datasets: ATG4D \cite{dpt} and nuScenes \cite{nuscenes2019}. 
We show that our approach can outperform the state-of-the-art  on both datasets. In particular, we significantly improve the social compliance between the estimated future trajectories, resulting in far fewer collisions between the predicted actors.
\end{abstract}

\section{Introduction}

Self-driving vehicles (SDVs) have the potential to change the way we live by providing safe, reliable and cost-effective transportation for everyone everywhere. 
In order to plan a safe maneuver, SDVs have to not only understand the past and the present but also predict the potential future for at least the duration of the motion planning horizon. 

Traditional software engineering stacks for self-driving are composed of a set of processes which are executed serially. Objects are first perceived in 3D via object detection to explain the current frame. These objects are then associated with detections from previous frames, and the trajectories are refined typically using a probabilistic filter.
The future positions and velocities of these objects are estimated by unrolling the state for the next few seconds, where a dynamical model (\eg, the bicycle model) is typically employed to produce physically realistic trajectories. An alternative approach is to associate an actor with a lane to form a goal (\eg, going straight, turning) and then predict its future trajectory towards that goal.
These models are, however, simplistic and have very little information as input. As a consequence, their estimates are often not very accurate.

More sophisticated approaches based on deep learning representations have been developed \cite{djuric2018motion,cui2018multimodal,bansal2018chauffeurnet} to increase the accuracy of the predictions. These approaches assume that the vehicle is localized and rasterize in bird's-eye view (BEV) both a high-definition  map (HD map) as well as the past and current detections. A convolutional neural network then learns to produce future trajectories from these rasters. 
However, all these approaches are slow as computation is not shared between the detection and motion forecasting networks. Instead, the forecasting network takes the raster images as input and performs many layers of computation. As a consequence real-time inference is very hard under this setting.
Moreover, as the prediction network does not have access to raw data, errors in the detection phase are very hard if not impossible to recover from due to information loss. 

In contrast, FAF \cite{dpt} developed a single network to jointly perform detection and motion forecasting. This results in better trajectories as the prediction module has access to the raw data. 
Furthermore, faster inference  is achieved as the computation is shared between tasks. 
IntentNet \cite{intentnet} proposed to output additionally a probability distribution over driving intentions (\eg, stopping, parked, lane change). 
\cite{neuralmp} took this a step further and jointly perform detection,  forecasting and motion planning. This resulted in the first end-to-end trainable motion planner that also produces interpretable intermediate representations.
While effective, all the aforementioned approaches ignore the statistical dependencies between actors, and instead predict each trajectory (or its probability) independently given the features. 

Various approaches have been proposed to model interactions between traffic actors and forecast motion given object detections. Ma et al. \cite{ma2017forecasting} formulated the interaction between pedestrians under a game-theoretic framework with fictitious play \cite{brown1951iterative}. \cite{sun2018courteous} introduced a courtesy term to an inverse reinforcement learning (IRL)  objective,  which better explains real-world human behavior. 
Other approaches model the interaction implicitly by message passing between actors \cite{alahi2016social,convsocial,ma2018trafficpredict}.
However, these methods use past ground truth trajectories of all actors as input. 
Self-driving cars do not have access to ground truth object trajectories and thus need to handle the uncertainty and noise of perception systems. Therefore, we tackle the end-to-end setting, i.e., from raw sensor data to detection and motion forecasting.

Towards this goal, we design a multi-sensor detector followed by an efficient interaction module that is inspired by the Transformer \cite{transformer}. We adapt the original architecture, which was developed for sequence modeling, to our problem domain by incorporating a novel pairwise attention mechanism with spatial reasoning via relative positions of the actors.
We also observe that the evolution of a trajectory largely depends on the behaviors of other actors in the recent past. To model this fact we propose a recurrent structure with per-time-step refinement to capture temporal dependencies between different prediction time steps.
Importantly, our architecture is very scalable since most of the computation scales 
linearly with the number of traffic actors. 
As a consequence, our end-to-end model runs at 10 FPS on a 1080Ti GPU.
We demonstrate the effectiveness of our approach on the challenging ATG4D \cite{dpt}  and  nuScenes\cite{nuscenes2019} datasets. Our experiments show that we outperform the state-of-the-art by a large margin in detection, trajectory accuracy and collision rate.

\section{Related Work}

In this section we review previous research on 3D object detection, trajectory prediction and  interaction modeling.

\subsubsection{3D Object Detection} 
Camera-based approaches take either monocular or stereo images \cite{mono3d,3dop,stereorcnn} as input. However, their performance is limited due to the difficulty of inferring depth. Recent  3D detectors rely on LiDAR. Different data representations have been developed: 3D voxels \cite{vote3deep}, bird's-eye view (BEV)  \cite{mv3d,pixor,hdnet,contfuse}, range view  \cite{mv3d,lasernet}, and point-wise operators \cite{voxelnet,fpointnet, pointrcnn}. Approaches that exploit multiple sensors \cite{mv3d,avod,fpointnet,contfuse,meyer2019sensor,mmf} achieve superior performance compared to single-sensor ones. Besides sensors, HD maps  provide useful priors \cite{hdnet,mmf}.

\subsubsection{Trajectory Prediction}  Dynamic models have been used to estimate the future states by propagating the current state over time \cite{cosgun2017towards,ziegler2014making}. However, these approaches are usually too simple to handle complexities in longer horizon prediction. Recent data-driven approaches have shown promising results in generating realistic future trajectories.
\cite{kim2017probabilistic} feeds the past vehicle locations in BEV into an LSTM network. \cite{djuric2018motion,cui2018multimodal,bansal2018chauffeurnet} generates BEV raster images that encode surrounding vehicle states and HD maps,  and then feeds them into a CNN. 
These approaches are however slow as the computation is not shared with the perception module.

\subsubsection{Joint Perception and Prediction} End-to-end models, on the contrary, are trained from raw sensor data and the perception and forecasting modules share features. Existing models \cite{dpt,intentnet,neuralmp} take multi-sweep LiDAR data and HD maps as input, and solve the two tasks using a single network. In addition, IntentNet \cite{intentnet} further reasons about semantic intentions.  NeuralMP \cite{neuralmp} further estimates jointly   motion planning. However, these methods do not explicitly take into account the interaction between actors, which plays an important role in real-world traffic.

\subsubsection{Interaction Modeling} Various frameworks have been proposed to model  multi-actor interactions, 
\eg, game theory \cite{ma2017forecasting}, Graph Neural Networks \cite{ma2018trafficpredict, spagnn}.
An alternative approach is feature aggregation between multiple actors. 
\cite{yi2016pedestrian} encodes the pedestrians in a 2D map and uses a CNN to capture the local interactions. \cite{alahi2016social,gupta2018social} propose a fully-connected layer (SocialPool) to aggregate information of neighboring pedestrians, while \cite{convsocial} further modifies the layer to be convolutional (ConvSocialPool). 
\cite{carnet,sophie,Huang_2019_ICCV,Choi_2019_ICCV, Kosaraju2019SocialBiGATMT, Ivanovic_2019_ICCV} propose explicit attention modules to identify important physical constraints and social neighbors.
In contrast, we propose to capture both temporal and spatial dependencies between actors by incorporating a Transformer-like \cite{transformer} module into a recurrent neural network. Towards this goal, we propose several important architecture modifications of the Transformer to handle our problem domain.
\section{Interaction Transformer}

\begin{figure*}[t]
\vspace{1mm}
\begin{center}
   \includegraphics[width=1.0\linewidth]{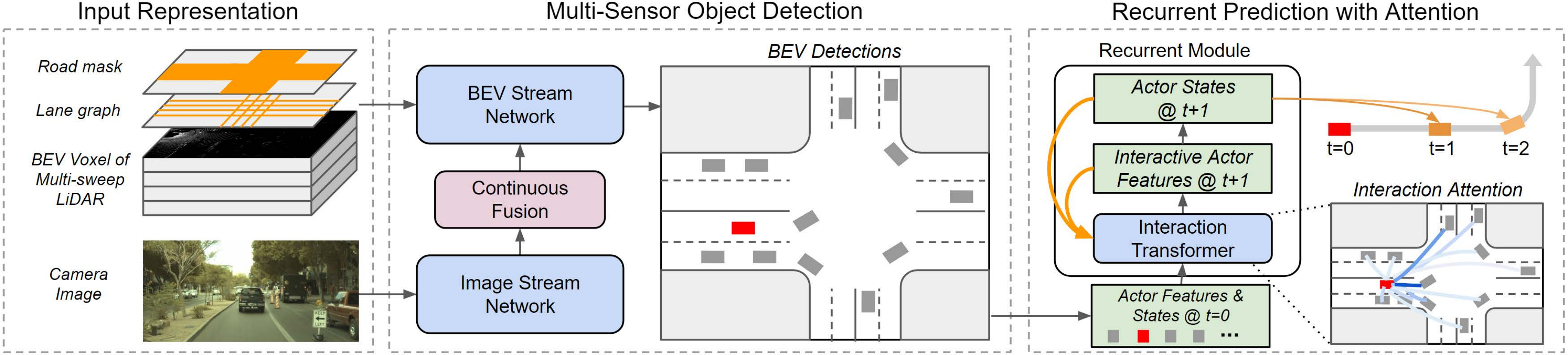}
\end{center}
\vspace{-3mm}
   \caption{\bd{Model architecture of our contextual end-to-end perception and prediction model.} We highlight one actor in the prediction module for clarity, while in practice all actors are predicted in parallel.}
\vspace{-3mm}
\label{fig:architecture}
\end{figure*}

In this paper, we propose an end-to-end model to solve the tasks of detection and motion forecasting from raw sensor data. Importantly, our model reasons about the interactions between the actors in the scene. 
We refer the reader to Fig. \ref{fig:architecture} for the overall model architecture. 
Our model has two main components: 1) a detection network that fuses information from multi-sensor data (i.e., LiDAR, images, HD maps) and outputs a set of bounding boxes, and 2) a recurrent interactive prediction module that processes interactions between actors and predicts their future trajectories.
In the subsequent sections, we first explain our model's input and output parameterizations, then describe each component of the model in detail, including our learning procedure.

\subsection{Input and Output Parameterizations}\label{sec:representation}

\subsubsection{Input Representation}
Given a LiDAR sweep, we voxelize the point cloud into a 3D occupancy grid in BEV with a fixed resolution, centered at the ego vehicle. We define each LiDAR point feature to be $1$, and compute the voxel feature by summing all nearby point features weighted by their relative positions to the voxel center. This helps preserve fine-grained information. 
To capture past motion for the task of motion forecasting, we aggregate multiple past LiDAR sweeps (registered to the current frame by ego-motion compensation) by concatenating their voxel representations along the $Z$ axis.
Following  \cite{hdnet}, we exploit both geometric and semantic map priors for better reasoning.
In particular, we subtract the ground height from the $Z$ value of each LiDAR point before voxelization. In this way, we can remove the variation caused by the ground slope.
We also extract semantic priors in the form of road and lane masks. 
Each of them is a one-channel raster image in BEV depicting the drivable surface and all the lanes respectively (see Fig. \ref{fig:architecture} left).
We augment the LiDAR voxel representation with semantic map priors by simple concatenation.

\subsubsection{Output Representation}
We define the output space in BEV as it enables efficient feature sharing between perception and forecasting. 
We parameterize the detection as a set of oriented bounding boxes. A detection box is represented with $(x, y, w, l, \theta)$, where $(x,y)$ is the box center, $(w,l)$ is the box size, and $\theta_i$ is its orientation. 
Note that the missing $Z$ dimension can be recovered from the ground prior in the HD map. 
Additionally, we represent the trajectory as a sequence of boxes at future  time steps, denoted as $\{(x_i^{(t)}, y_i^{(t)}, \theta_i^{(t)})\}$, with $t = 1, ..., T$. We assume that the objects are rigid and thus their sizes are kept the same across all time steps. Below we use the terms of trajectory prediction and motion forecasting interchangeably.

\subsection{Object Detection from Multi-Sensors}\label{sec:detection}
\subsubsection{Backbone Network Architecture}
The sensor-fusion backbone network follows the two-stream architecture of \cite{mmf}.
The BEV stream is a customized 2D CNN that extracts features in BEV space from joint LiDAR and map representation. Inception-like blocks are stacked sequentially with residual connections to extract multi-scale feature maps. 
The image stream is a ResNet-18 \cite{residual} pre-trained on ImageNet \cite{imagenet}, and receives RGB camera images as input. 
We aggregate multi-scale image feature maps from each ResNet-18 residual block with a feature pyramid network \cite{fpn}, and fuse the aggregated feature map with the BEV stream via a continuous fusion layer \cite{contfuse}, providing dense fusion.
Specifically, image features are first back-projected to BEV space according to the existing LiDAR observation. At BEV locations with no LiDAR points, the image features are interpolated from nearby occupied locations via an MLP. The image and  BEV features are then fused by element-wise addition in BEV space.
The output feature map from the BEV stream is used to provide multi-sensor features for the detection and prediction modules. 

\subsubsection{Dense BEV Object Detection}
As vehicles are relatively similar and do not overlap in BEV space, following RetinaNet \cite{focal}, we formulate object detection as dense prediction without any object anchors.
We apply several $1\times 1$ convolutions on top of the BEV feature map, which outputs an 8-channel vector per voxel, representing a confidence score $s$ and a bounding box parameterized as $(dx, dy, w, l, \sin2\theta, \cos2\theta, cls_\theta)$, where $(dx,dy)$ are the relative position offsets from the voxel center to the box center, $(w,l)$ are the box size, and $(\sin2\theta, \cos2\theta, cls_\theta)$ are used to decode the orientation.
Following \cite{hdnet}, we regress to $(\sin2\theta, \cos2\theta)$ to get the orientation without distinguishing  flips. To effectively model interaction we need to understand which direction each vehicle is facing.
We use an additional classification step $cls_\theta$ to classify whether the orientation is in ($\frac{1}{4}\pi$, $\frac{3}{4}\pi$]~\cite{2019arXiv190809492Z}. This classification target is better than (0, $\pi$] since orientations of many vehicles are similar or opposite to the orientation of the ego vehicle, which are at the boundaries of (0, $\pi$]. We finally apply oriented non-maximum suppression (NMS) to remove the duplicates and keep all remaining boxes whose score is above a threshold. 

\subsection{Recurrent Interactive Motion Forecasting}\label{sec:prediction}
We based our design on two observations.
First, the behaviors of actors heavily depend on each other. For example, drivers control the vehicle speed to keep a safe distance to the vehicle ahead.  At intersections, drivers typically wait for those that have the right of way.
Second,  the output at each time step depends on the outputs at previous time steps.
We thus propose a recurrent interactive motion forecasting model that 1) jointly reasons about all actors modeling their interactions and 2) iteratively infers the trajectory to capture its sequential nature.
Our interaction module is inspired by the {\it Transformer} \cite{transformer}, an 
architecture developed  for machine translation. We adapt it to our motion forecasting task. 
Below we first review the Transformer module, then describe the differences with our novel {\it Interaction Transformer} module, and finally explain the recurrent inference process.

\subsubsection{A Review on the Transformer}
Sequence to sequence models with an encoder-decoder architecture have been predominant in natural language processing. In this context, the {\it Transformer} \cite{transformer} was proposed as an attention mechanism that can be used to draw global dependencies between inputs and outputs, especially for long sequences. 
The Transformer projects each feature to a query and a key-value pair, which are all vectors. For each query, it computes a set of attentional weights using a compatibility function between the query and the set of keys. The output feature is then the sum of values weighted by the attention, plus some nonlinear transformations.

More formally, we denote the input sequence as $F^{in} \in \mathbb{R}^{n \times d_f}$, each row of which is a feature vector. The Transformer uses linear projections to get the set of queries, keys and values as follows
\begin{align}
Q = F^{in}W^Q, \,\,\, K = F^{in}W^K, \,\,\, V = F^{in}W^V,
\label{equ:projection}
\end{align}
where $W^Q \in \mathbb{R}^{d_f \times d_k}$, $W^K \in \mathbb{R}^{d_f \times d_k}$ and $W^V \in \mathbb{R}^{d_f \times d_v}$ are matrices of weights.
It then uses scaled dot products between the queries and keys to compute the attentional weights, and then aggregates the values for each query
\begin{align}
A = \textrm{softmax}\left(\frac{QK^T}{\sqrt{d_k}}\right)V,
\label{equ:attention}
\end{align}
where a softmax function is used to add a sum-to-one normalization to the attentional weights of a query (each row of $QK^T$). 
The scaling factor $\frac{1}{\sqrt{d_k}}$ is used to prevent the dot product from being numerically too large.
Finally, the Transformer uses a set of non-linear transformations with shortcut connections to get the output features:
\begin{align}
F^{out} = \textrm{ResBlock}(\textrm{MLP}(A) + F_{in}),
\label{equ:nonlinear}
\end{align}
where MLP is a Multi-Layer Perception applied to each row of $A$, and ResBlock is a residual block \cite{residual} also applied on each row. The output $F^{out}$ has the same shape as $F^{in}$.

\begin{figure*}[t]
\vspace{1mm}
\begin{center}
   \includegraphics[width=\linewidth]{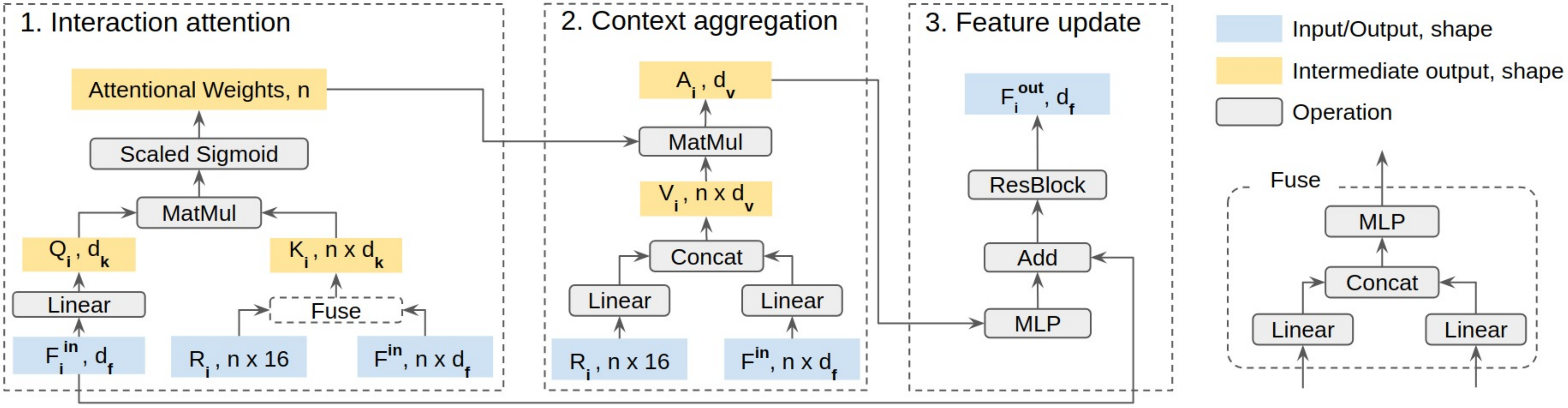}
\end{center}
\vspace{-3mm}
\caption{\bd{Interaction Transformer module.}
The figure shows the computation of contextual feature for actor $i$. We use actor $i$'s feature $F_i^{in}$ to compute the query. We use all $n$ actors' features $F^{in}$ and actor $i$'s location embedding $R_i$ relative to all actors to compute the keys and values. We use the query and keys to obtain the attentional weights, which are then used to aggregate the values to $F_i^{in}$. We apply a residual block to get the updated contextual feature $F_i^{out}$ for actor $i$.}
\vspace{-3mm}
\label{fig:transformer}
\end{figure*}

\subsubsection{Our Interaction Transformer}
The Transformer was designed to process an input sequence. In our task, the input is a set of actors and their representations. We represent the state of each actor with features extracted from the BEV feature map as well as the actor's spatial information, which contains the center location, size, and orientation of the actor.
In our task, the queries Q represent the actors we want to forecast their motion of, and keys K and values V represent neighbor contextual information from other actors. 
Fig. \ref{fig:transformer} shows  a diagram of our Interaction Transformer.

We denote the input actor features as $F^{in}$, where each row is the feature vector of an actor. 
We use Eq. (\ref{equ:projection}) to compute the queries $Q$.
However, there are significant differences in computing the keys and values due to the spatial nature of our problem. 
In the Transformer, location information is encoded as \emph{absolute} positional embeddings which are fused into the input features. 
In our task, absolute locations provide little information about the interactions between actors,
yet \emph{relative} locations and orientations are crucial.
Thus for each target actor $i$, we transform other actors into the local coordinate system centered on the $i$-th actor with the x-axis aligned with the orientation of the actor. 
This transformation ensures learning of the interaction to be translation and rotation invariant, which results in better performance.
We then encode the relative location information with the relative location embedding $R$, where $R \in \mathbb{R}^{n \times n \times 16}$ is a 3-dimensional matrix:
\begin{align}
R_{ij} = \textrm{MLP}( | x_j^i |, | y_j^i |, \textrm{sgn}(x_j^i), \textrm{sgn}(y_j^i), w_j, l_j, \theta_j^i).
\end{align}
$x_j^i$, $y_j^i$, and $\theta_j^i$ are the locations and orientation of actor $j$ after it has been transformed into actor $i$'s local coordinate system,  $sgn$ is the sign function, and $(w_j,l_j)$ encode the actor's size. A two-layer MLP transforms the 7-channel input to the 16-channel embedding.

We use actor features $F^{in}$ and the relative location embedding $R$ to compute $K$ and $V$. As $R$ cannot be fused into the single actor feature, we cannot use a shared set of keys and values for different queries. In this case, both $K$ and $V$ become 3-dimensional matrices, i.e., $K \in \mathbb{R}^{n \times n \times d_k}$ and $V \in \mathbb{R}^{n \times n \times d_v}$. We compute $K$ and $V$ using:
\begin{align}
&K_i = \textrm{MLP}(\textrm{Concat}(F^{in}W^{K_1}, R_i W^{K_2})), \\
&V_i = \textrm{Concat}(F^{in}W^{V_1}, R_i W^{V_2}).
\end{align}
where Concat denotes concatenation along the second dimension, and an MLP is applied on each row vector.
We then compute the attention as:
\begin{align}
A_i = \textrm{sigmoid}\left(\frac{Q_iK_i^T}{\sqrt{d_k}}\right)V_i.
\end{align}
Note that we change the softmax function in  Eq.~(\ref{equ:attention}) to a sigmoid function. For each prediction target, the softmax function forces the attention scores of all other actors to sum to 1, which is not a reasonable assumption in our context, as this 
sum is expected to be low when the target actor is not interacting with others, and high when the target actor has strong interactions.
Finally, we compute the output features following Eq. (\ref{equ:nonlinear}).

\subsubsection{Recurrent Temporal Prediction}
To capture the sequential nature of the trajectory outputs, we use a recurrent model to predict the motion in an autoregressive fashion (see Fig. \ref{fig:architecture} right), where the number of recurrent steps equals the number of predicted time steps.
At time step $t$, the relative location embedding $R^{(t)}$ is computed from the output waypoints at $t-1$, and the input actor feature is denoted as $F^{in (t)}$.
We use detection bounding boxes to compute $R^{(0)}$, and set $F^{in (0)}$ to be the bi-linearly interpolated output BEV features extracted at the detection box centers. 
The Interaction Transformer takes $F^{in (t)}$ and $R^{(t)}$ as input and outputs $F^{out (t)}$, which is then fed into a two-layer MLP to compute the next set of output waypoints. We then  use $F^{out (t)}$ as $F^{in (t + 1)}$.

\subsubsection{Per-Time-Step refinement}
For each time step, we use an extra refinement step to obtain more accurate waypoints. 
The Interaction Transformer first takes $F_{\text{in}}^{t-1}$ and relative location embedding $R^{t-1}$ to compute $F_{\text{out}}^{t-1}$ and predict a waypoint proposal $P_{\text{proposal}}^{t}$. 
The updated feature $F_{\text{in}}^{t}$ (same as $F_{\text{out}}^{t-1}$) and $R^{t}$ (computed from $P_{\text{proposal}}^{t}$) are then fed into the Interactive Transformer again to obtain the refined waypoints $P_{\text{refine}}^{t}$ as our final motion forecasting output. 
Compared to the proposal step, the refinement step encodes more up-to-date spatial relations between actors. 
For computational efficiency, $P_{\text{refine}}$ and $P_{\text{proposal}}$ are computed in parallel by a single forward pass of the Interaction Transformer. 
\ie, we obtain both $P_{\text{refine}}^{t}$ and $P_{\text{proposal}}^{t+1}$ from $F_{\text{out}}^{t}$, by two different MLPs.
Both $P_{\text{refine}}$ and $P_{\text{proposal}}$ are parameterized as $(dx, dy, \sin2\theta, \cos2\theta, cls_\theta)$, where ($dx$, $dy$) are relative to the detection box center, and $(\sin2\theta, \cos2\theta, cls_\theta)$ are defined in the same way as the detection output parameterizations. 

\subsection{Learning}\label{sec:learning}
Our model is fully differentiable and can be trained end-to-end by minimizing the weighted sum of detection and prediction losses
$$L_{\text{total}} = L_{\text{detect}} + \alpha (L_{\text{predict}}^{\text{proposal}} + L_{\text{predict}}^{\text{refine}}),$$
where $\alpha$ is  a scalar. For detection loss we use the weighted sum of  three losses: object classification loss, $\ell_2$ loss for the  box regression, and orientation classification loss for the orientation flip.  
Similarly, we use the sum of three losses as our motion forecasting loss: the $\ell_2$ loss for the trajectory and future heading,  the classification loss for the orientation flip, and a collision loss that computes the negative $\ell_2$ distance between two vehicles only in the case that there is a collision. This collision loss pushes colliding vehicles away from each other.

For object detection, we use the distance between BEV voxels and their closest ground-truth box center to determine positive and negative samples. 
Samples whose distance is smaller than a threshold are considered as positive, and negative otherwise. As a large proportion of the samples are negative in dense object detection, we adopt online hard negative mining, where we only keep the most difficult negative samples (with the largest loss) and ignore the easy negative ones.
Classification loss is computed over both positive and negative samples while regression loss is computed over positive samples only.

We perform online associations between detection results and ground-truth labels to compute the motion forecast loss. 
For each detection, we assign the ground-truth box with the maximum (oriented) IoU.
If a ground truth box is assigned to multiple detections, only the detection with maximum IoU is kept while other detections are ignored. 
Regression on future motion is then averaged over those detections with associated ground-truth.
Compared to regression on all ground truth actors, our motion forecast loss will not be dominated by actors which are not detected.

\section{Experimental Evaluation}

\subsubsection{Datasets}
We evaluate our approach on two large-scale real-world driving datasets: ATG4D \cite{dpt} and nuScenes\cite{nuscenes2019}.
\textbf{ATG4D} was collected by driving in multiple North American cities with a multi-sensor kit mounted on top of a fleet of vehicles. In total it contains 5,500 video snippets of 25 seconds each with a frame sampling rate of 10 Hz.
We use 5,000 snippets for training and 500 snippets for evaluation. 
Although scenario diversity has already been considered when creating the ATG4D dataset, interactions between traffic actors still happen relatively rarely. 
To address this problem, we also sample an additional evaluation subset \textbf{ATG4D-interact} from ATG4D-eval which contains interacting actors. Specifically, we focus on blocking interactions by defining actors that: 1) are not parked 2) have a velocity smaller than 0.2 m/s; 3) and have another vehicle nearby ($<$5m) in front and in the same lane. We search in the whole dataset for keyframes when there is a vehicle entering or leaving the blocking status. We extend each keyframe temporally by 5 seconds from the past or into the future depending on whether the actor is entering or leaving the blocking status. Using these heuristics, we collect around 800 5-second snippets that are considered “interactive”.
\textbf{nuScenes} contains 850 scenes of 20 seconds of driving data, with a frame sampling rate of 20 Hz. Positions of all actors are labeled at 2Hz. We follow the official splits.


\begin{table*}[t]
\vspace{2mm}
\caption\rightmark\vspace{-4mm}
{\bd{Comparison with state-of-the-art on ATG4D-eval testset.} \emph{AP} means Average Precision as object detection metric. \emph{ADE} and \emph{FDE} mean Average Displacement Error and Final Displacement Error as trajectory prediction metrics. \emph{TCR} means Trajectory Collision Rate that measures social compliance between actors. $\uparrow$ means the higher the better, $\downarrow$ means the lower the better.}
\begin{center}
\begin{tabular}{l||cc|cc|cc|cc}
\shline
\multirow{2}{*}{Method} & \multicolumn{2}{c|}{AP (\%) $\uparrow$} & \multicolumn{2}{c|}{ADE (cm) $\downarrow$} & \multicolumn{2}{c|}{FDE (cm) $\downarrow$} & \multicolumn{2}{c}{TCR(\%) $\downarrow$}\\
 & IoU0.5 & IoU0.7 & recall0.7 & recall0.9 & recall0.7 & recall0.9 & recall0.7 & recall0.9 \\
\shline
FAF \cite{dpt} & 90.20 & 72.93 & 72.8 & 82.5 & 131.1 & 145.2 & 0.729 & 1.242 \\
NeuralMP \cite{neuralmp} & 89.83 & 76.15 & 63.1 & 74.4 & 114.7 & 129.0 & 0.408 & 0.656 \\
\hline
SocialPool \cite{alahi2016social} & 94.60 & 82.02 & 55.4 & 63.8 & 97.7 & 109.2 & 0.288 & 0.403 \\
ConvSocialPool \cite{convsocial} & 94.47 & 81.98 & 55.0 & 63.5 & 96.5 & 108.1 & 0.207 & 0.310 \\
\hline
Ours, no collision loss & 94.62 & \bd{82.18} & \bd{52.6} & \bd{61.1} & 92.1 & \bd{104.1} & 0.080 & 0.126\\
Ours, with collision loss & \bd{94.68} & 82.09 & \bd{52.6} & 61.3 & \bd{92.0} & 104.4 & \bd{0.041} & \bd{0.055}\\
\shline
\end{tabular}
\end{center}
\vspace{-2mm}
\label{tab:result}
\end{table*}

\begin{table*}[t]
\caption\rightmark\vspace{-4mm}
{\bd{Comparison with state of the art on nuScenes validation set.} We follow the same evaluation metrics as SpAGNN\cite{spagnn}, which reports L2 error of actor's center, at 0s, 1s, and 3s, and uses IoU threshold 0.1 for NMS and collision rate. They use IoU threshold 0.5 to associate detections with ground truth, and shows results with recall threshold of 60\%. Since recall is lower under this high-association-IoU setting, we show additional results at recall 80\% instead of 90\%.}
\begin{center}
\footnotesize
\begin{tabular}{l||cc|cc|cc|cc|cc}
\shline
\multirow{2}{*}{Method} & \multicolumn{2}{c|}{AP (\%)$\uparrow$} & \multicolumn{2}{c|}{L2 @ 0s (cm) $\downarrow$} & \multicolumn{2}{c|}{L2 @ 1s (cm) $\downarrow$} & \multicolumn{2}{c|}{L2 @ 3s (cm) $\downarrow$} & \multicolumn{2}{c}{TCR IoU0.1 (\%)$\downarrow$}\\
 & IoU0.5 & IoU0.7 & recall0.6 & recall0.8 & recall0.6 & recall0.8 & recall0.6 & recall0.8 & recall0.6 & recall0.8 \\
\shline
SpAGNN \cite{spagnn} & - & - & 22 & - & 58 & - & 145 & - & 0.222 & - \\
\hline
SocialPool \cite{alahi2016social} & 82.53 & 70.11 & 20.1 & 23.1 & 39.2 & 44.0 & 119.9 & 127.6 & 0.158 & 0.881 \\
ConvSocialPool \cite{convsocial} & 82.16 & 69.55 & 21.2 & 24.2 & 39.3 & 44.2 & 117.6 & 124.5 & 0.174 & 0.722 \\
\hline
Ours, no collision loss & \bd{82.72} & \bd{70.29} & \bd{19.6} & \bd{22.5} & \bd{38.2} & \bd{42.7} & 112.7 & 118.3 & 0.057 & 0.269 \\
Ours, with collision loss & 82.66 & 70.12 & 20.1 & 22.9 & 38.4 & 43.0 & \bd{112.4} & \bd{117.9} & \bd{0.016} & \bd{0.058}\\
\shline
\end{tabular}
\end{center}
\vspace{-2mm}
\label{tab:result_nuscenes}
\end{table*}
 

\begin{table*}[t]
\begin{center}
\caption\rightmark\vspace{-2mm}
{\bd{Ablation study of motion prediction architecture and collision loss.}\vspace{2mm}}
\footnotesize
\setlength\tabcolsep{3pt}
\scalebox{0.92}{
\begin{tabular}{c|c|c|c|c||c|c|c|c||c|c|c|c}
\shline
\multirow{3}{3.9em}{2nd-stage prediction} & \multirow{3}{3.3em}{collision loss} & \multirow{3}{*}{Transformer} & \multirow{3}{*}{recurrent} & \multirow{3}{*}{refine} & \multicolumn{4}{c||}{ATG4D-interact} & \multicolumn{4}{c}{nuScenes} \\
\cline{6-13}
&&&&& AP (\%) $\uparrow$ & ADE (cm) $\downarrow$ & FDE (cm) $\downarrow$ & TCR (\%) $\downarrow$  & AP (\%)  $\uparrow$ & ADE (cm) $\downarrow$ & FDE (cm) $\downarrow$ & TCR (\%) $\downarrow$ \\
&&&&& IoU0.7 & recall0.9 & recall0.9 & recall0.9 & IoU0.5 & recall0.9 & recall0.9 & recall0.9 \\
\shline
\textcolor{mygray}{\xmark} & \textcolor{mygray}{\xmark} & \textcolor{mygray}{\xmark} & \textcolor{mygray}{\xmark} & \textcolor{mygray}{\xmark} & 79.13 & 89.8 & 159.6 & 2.094 & 81.43 & 80.5 & 134.5 & 2.004 \\
\cmark & \textcolor{mygray}{\xmark} & \textcolor{mygray}{\xmark} & \textcolor{mygray}{\xmark} & \textcolor{mygray}{\xmark} & 80.31 & 87.9 & 157.7 & 1.893 & 82.45 & 80.3 & 135.1 & 2.052 \\
\cmark & \cmark & \textcolor{mygray}{\xmark} & \textcolor{mygray}{\xmark} & \textcolor{mygray}{\xmark} & 80.40 & 89.7 & 159.9 & 1.701 & 82.55 & 80.8 & 135.2 & 1.667 \\
\cline{1-13}
\cmark & \textcolor{mygray}{\xmark} & \cmark & \textcolor{mygray}{\xmark} & \textcolor{mygray}{\xmark} & 80.44 & 85.1 & 152.4 & 0.885 & \bd{82.88} & 78.4 & 130.1 & 1.314 \\
\cmark & \textcolor{mygray}{\xmark} & \cmark & \cmark & \textcolor{mygray}{\xmark} & \bd{80.51} & 83.9 & 149.3 & 0.529 & 82.76 & 77.0 & 126.0 & 0.941 \\
\cmark & \textcolor{mygray}{\xmark} & \cmark & \cmark & \cmark & 80.46 & \bd{83.7} & \bd{147.4} & 0.354 & 82.72 & \bd{76.6} & 125.2 & 0.909 \\
\cmark & \cmark & \cmark & \cmark & \cmark & 80.37 & 84.0 & 147.9 & \bd{0.157} & 82.66 & 76.8 & \bd{124.9} & \bd{0.122} \\
\shline
\end{tabular}}
\end{center}
\label{tab:ablation1}
\vspace{-2mm}
\end{table*}

\begin{table*}[t]
\begin{center}
\caption\rightmark\vspace{-2mm}
{\bd{Ablation study of the Interaction Transformer.}\vspace{2mm}}
\footnotesize
\scalebox{0.92}{
\begin{tabular}{c|c|c||c|c|c|c||c|c|c|c}
\shline
\multirow{3}{2.5em}{actor feature} & \multirow{3}{3.9em}{relative location embedding} & \multirow{3}{5em}{attention normalization} & \multicolumn{4}{c||}{ATG4D-interact} & \multicolumn{4}{c}{nuScenes} \\
\cline{4-11}
&&& AP (\%) $\uparrow$ & ADE (cm) $\downarrow$ & FDE (cm) $\downarrow$ & TCR (\%) $\downarrow$  & AP (\%)  $\uparrow$ & ADE (cm) $\downarrow$ & FDE (cm) $\downarrow$ & TCR (\%) $\downarrow$ \\
&&&IoU0.7  & recall0.9 & recall0.9 & recall0.9  & IoU0.5 & recall0.9 & recall0.9 & recall0.9 \\
\shline
\cmark & \textcolor{mygray}{\xmark} & softmax & 80.40 & 86.9 & 155.1 & 1.439 & 82.56 & 79.9 & 134.1 & 1.923 \\
\cmark & global  & softmax & 80.33 & 85.0 & 150.8 & 0.853 & 82.88 & 77.9 & 127.1 & 1.137 \\
\cmark & actor  & softmax & 80.27 & 84.6 & 149.0 & 0.476 & \bd{82.94} & 77.5 & 126.3 & 1.043 \\
\cmark & actor  & sigmoid & \bd{80.46} & \bd{83.7} & \bd{147.4} & \bd{0.354} & 82.72 & \bd{76.6} & \bd{125.2} & \bd{0.909} \\
\shline
\end{tabular}}
\end{center}
\label{tab:ablation2}
\vspace{-6mm}
\end{table*}

\subsubsection{Evaluation Metrics}
{\it Detection Average Precision (AP)} is defined as the area under the IoU based Precision-Recall curve. Note that AP measures the performance on the object detection task alone. 
For a comprehensive analysis, we compute AP at  $0.5$ and $0.7$ IoUs respectively.
{\it Average Displacement Error (ADE)} is defined as the mean of $\ell_2$ distances between the predicted future trajectory and the ground-truth trajectory at all time steps within the prediction horizon. 
{\it Final Displacement Error (FDE)} is defined as the $\ell_2$ distance between the predicted and ground-truth trajectories at the end of the prediction horizon. 
{\it Trajectory Collision Rate (TCR)} is defined as the collision rate between the future trajectories of object detections, where collision is defined as $>0.05$ IoU between two detections at any time-step within the prediction horizon.
ADE measures the overall accuracy of trajectory prediction, 
whereas FDE highlights the performance in long-term prediction.
TCR measures the social compliance of the actors and therefore is used to evaluate the effectiveness of the proposed interaction module. 
ADE, FDE, and TCR are evaluated on all vehicles, regardless of whether they are moving or not. 
For a fair comparison between models with different detections, we compute ADE, FDE, and TCR at a fixed 70\% and 90\% recall rates (with 0.1 IoU threshold) respectively.

\subsubsection{Implementation Details}
We use a voxel resolution of $0.156$m for the input BEV representation. We aggregate the past 5 LiDAR sweeps for ATG4D and 10 sweeps for nuScenes to encode a 0.5-second history.
We train our model end-to-end using the Adam optimizer \cite{adam}. 
We ignore object labels that are not observable by the LiDAR (\ie, $0$ points inside the box).
We apply NMS with 0.05 IoU threshold to the detection results to avoid object collision at $t=0$. We also remove detections with $<0.1$ confidence score. The prediction horizon of our model is 3 seconds, with a time-step interval of 0.5 second (therefore $T=6$).
For ATG4D, we train and evaluate the model in the front region of the ego-car within a 100m range. For nuScenes, we follow the official evaluation range, which is 50m around the ego-car.
We do not use images or maps in nuScenes because not every frame has well-aligned images and some map data has large alignment error with LiDAR.
Due to the limited number of labels in nuScenes, we also conduct augmentation during training. 
Since labels are only available at 2Hz,  we use linear interpolation to estimate actors' bounding boxes for frames without labels. 
We also conduct spatial augmentation by randomly scaling (uniformly sampling  from [0.95, 1.05]), rotating  the ground plane between $\left[-\frac{\pi}{6}, \frac{\pi}{6}\right]$, and translating in the range [1.0, 1.0, 0.2] m for x, y, and z axes respectively.

\begin{figure*}[t]
\begin{center}
\includegraphics[width=0.195\linewidth,trim={0 0 0 10cm},clip]{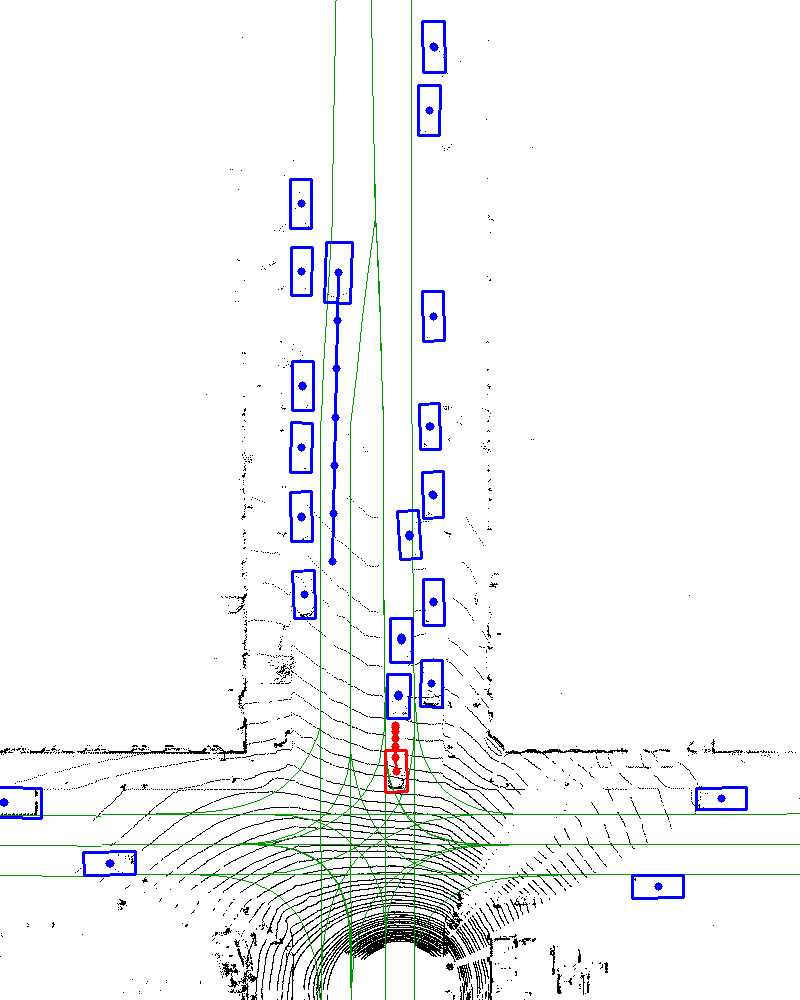}
\includegraphics[width=0.195\linewidth,trim={0 10cm 0 0},clip]{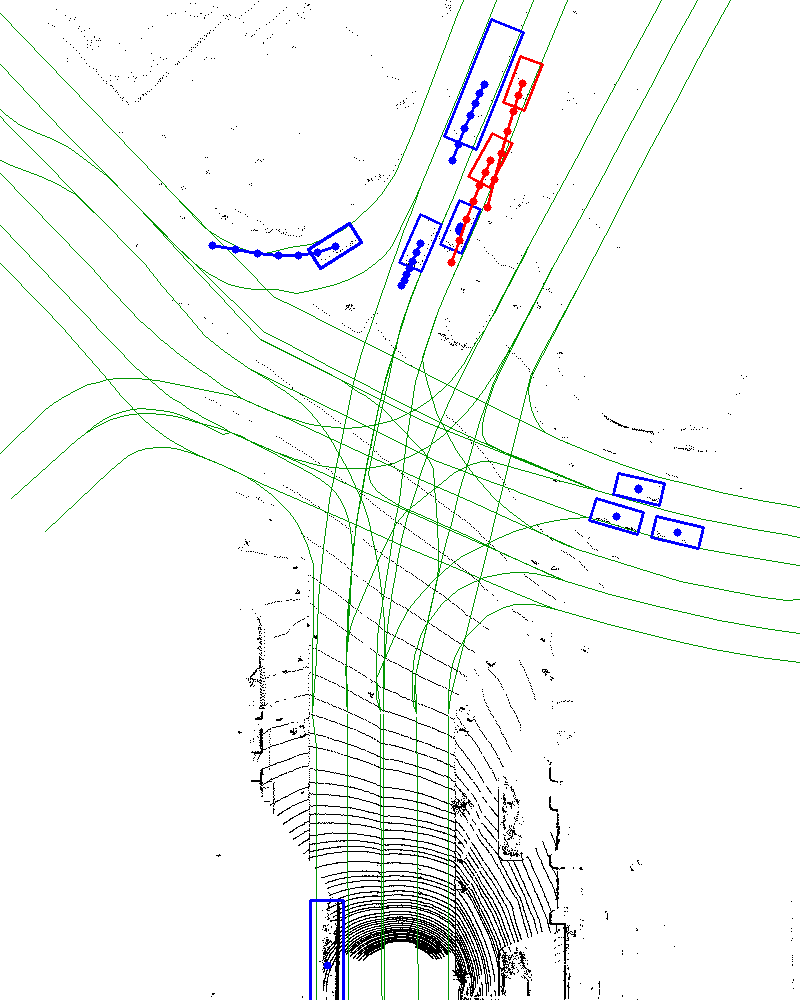}
\includegraphics[width=0.195\linewidth,trim={0 0 0 10cm},clip]{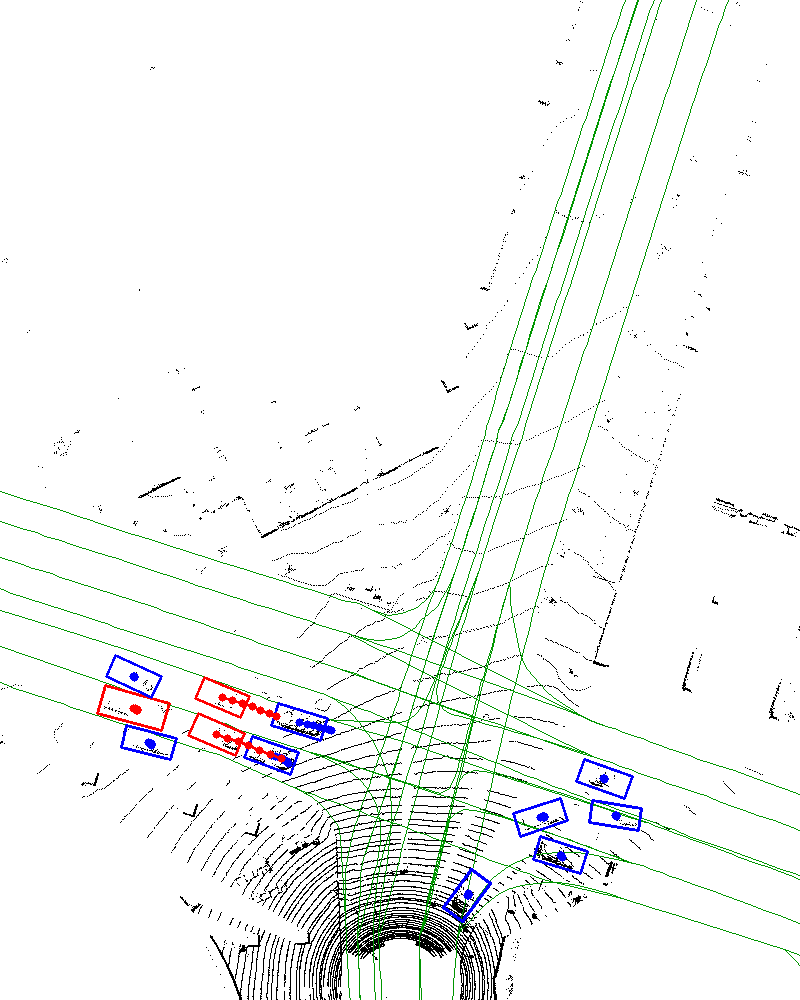}
\includegraphics[width=0.195\linewidth,trim={0 0 0 10cm},clip]{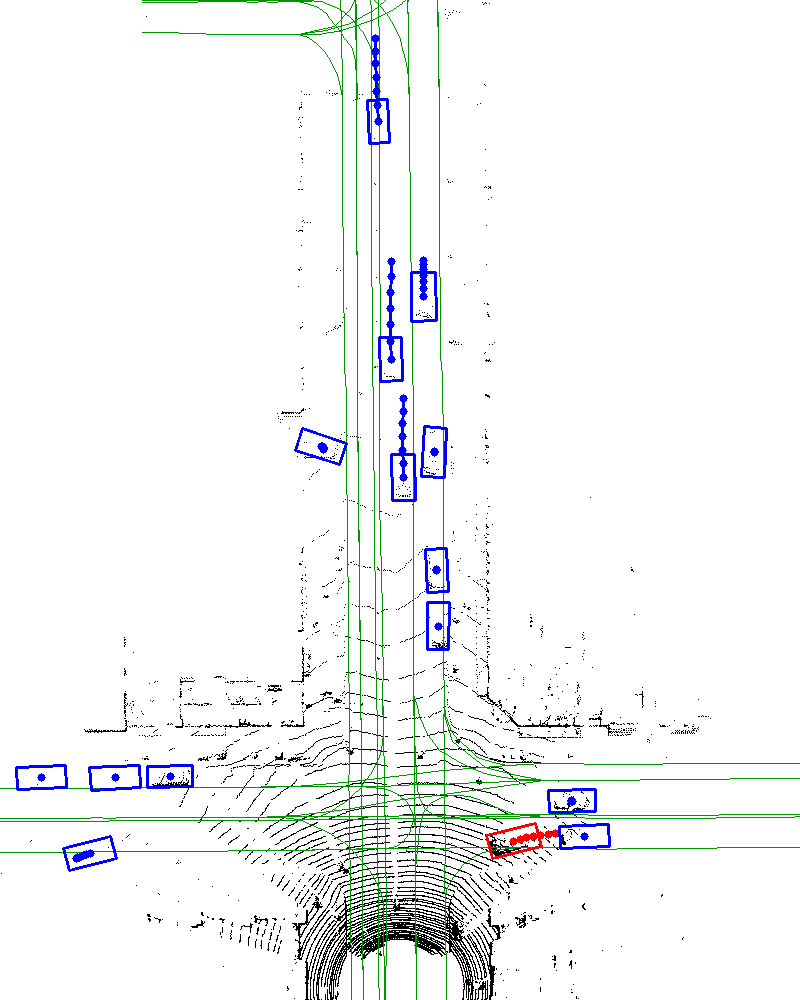}
\includegraphics[width=0.195\linewidth,trim={0 10cm 0 0},clip]{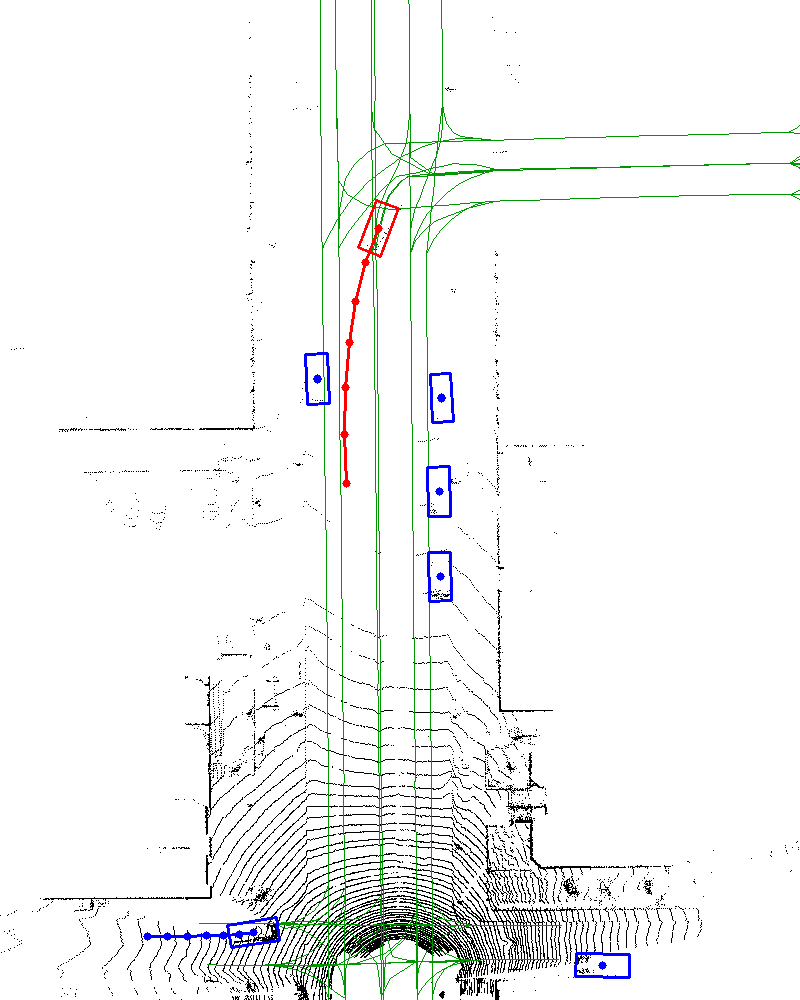}\\
\includegraphics[width=0.195\linewidth,trim={0 0 0 10cm},clip]{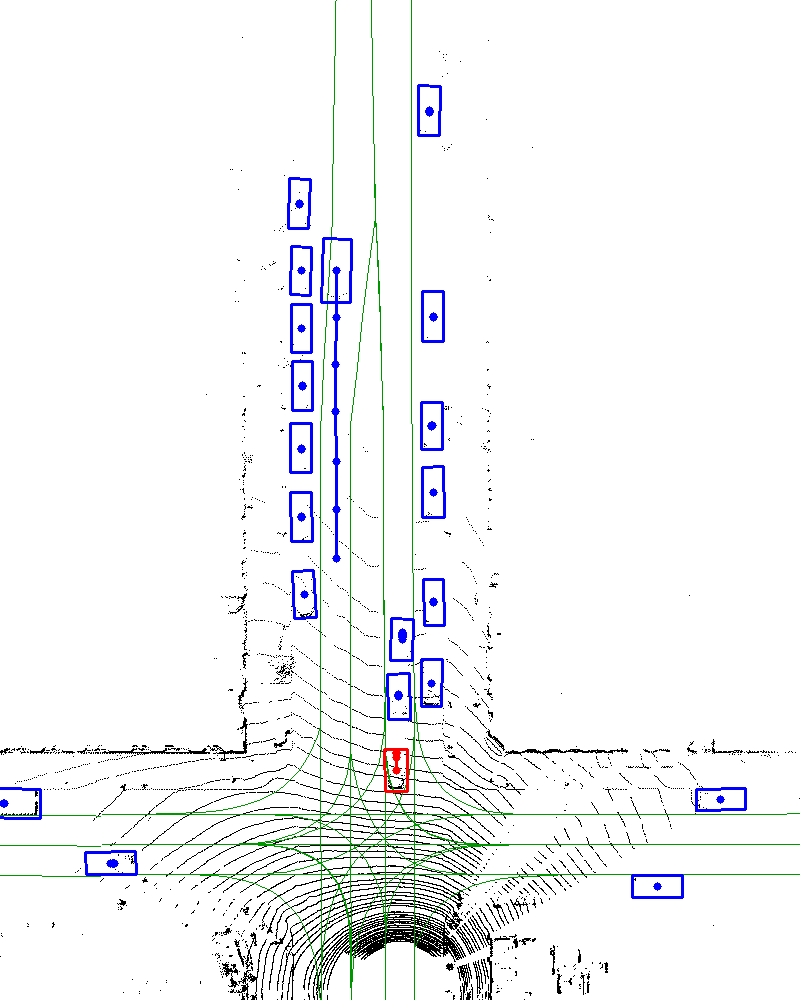}
\includegraphics[width=0.195\linewidth,trim={0 10cm 0 0},clip]{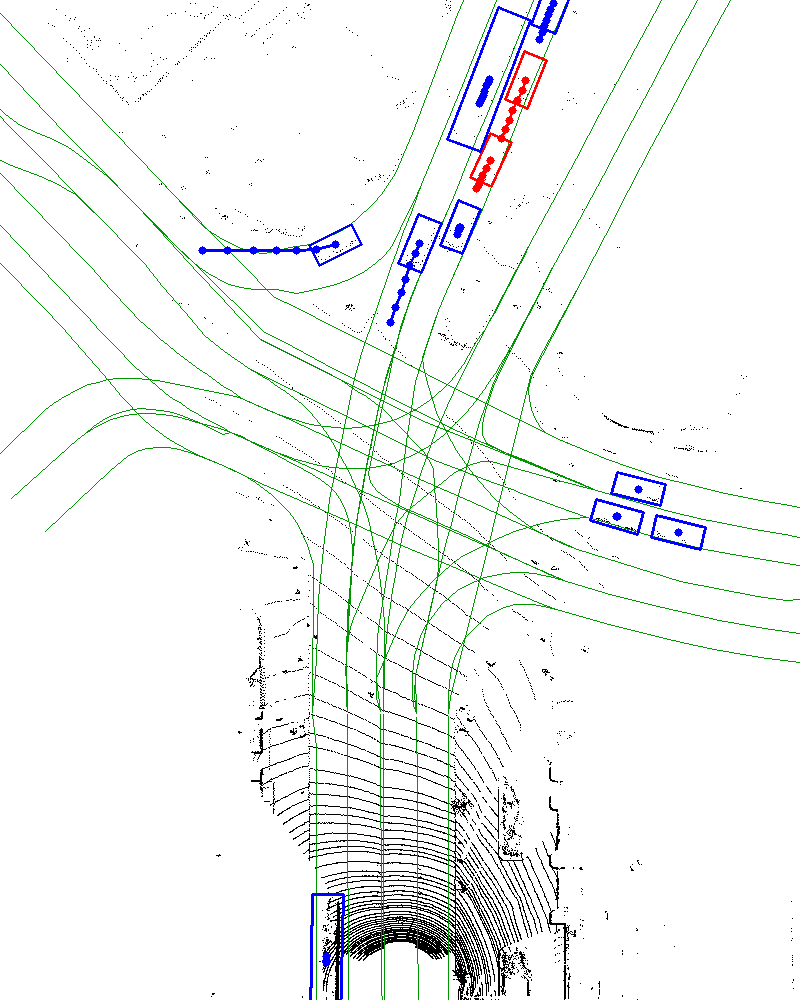}
\includegraphics[width=0.195\linewidth,trim={0 0 0 10cm},clip]{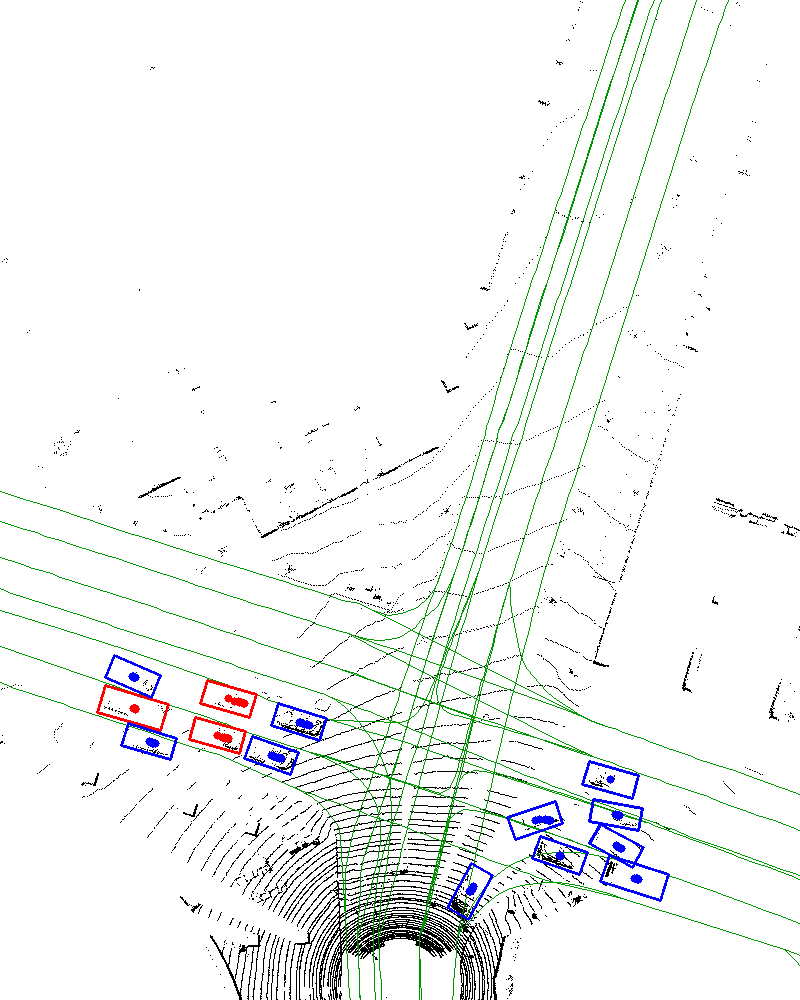}
\includegraphics[width=0.195\linewidth,trim={0 0 0 10cm},clip]{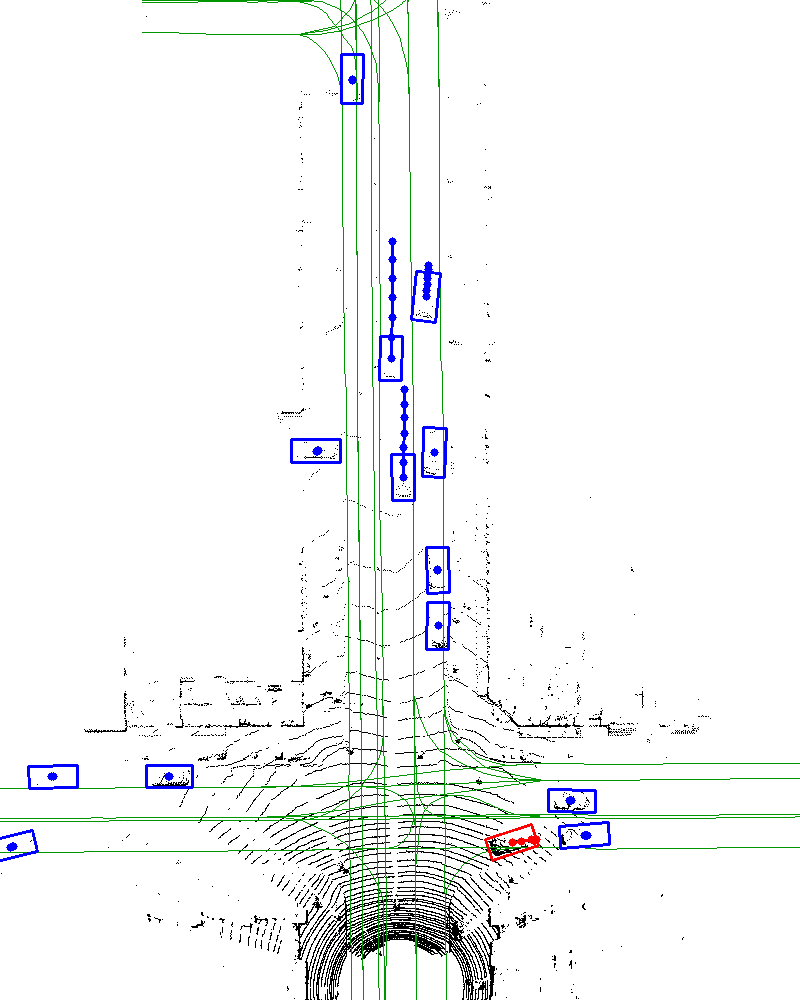}
\includegraphics[width=0.195\linewidth,trim={0 10cm 0 0},clip]{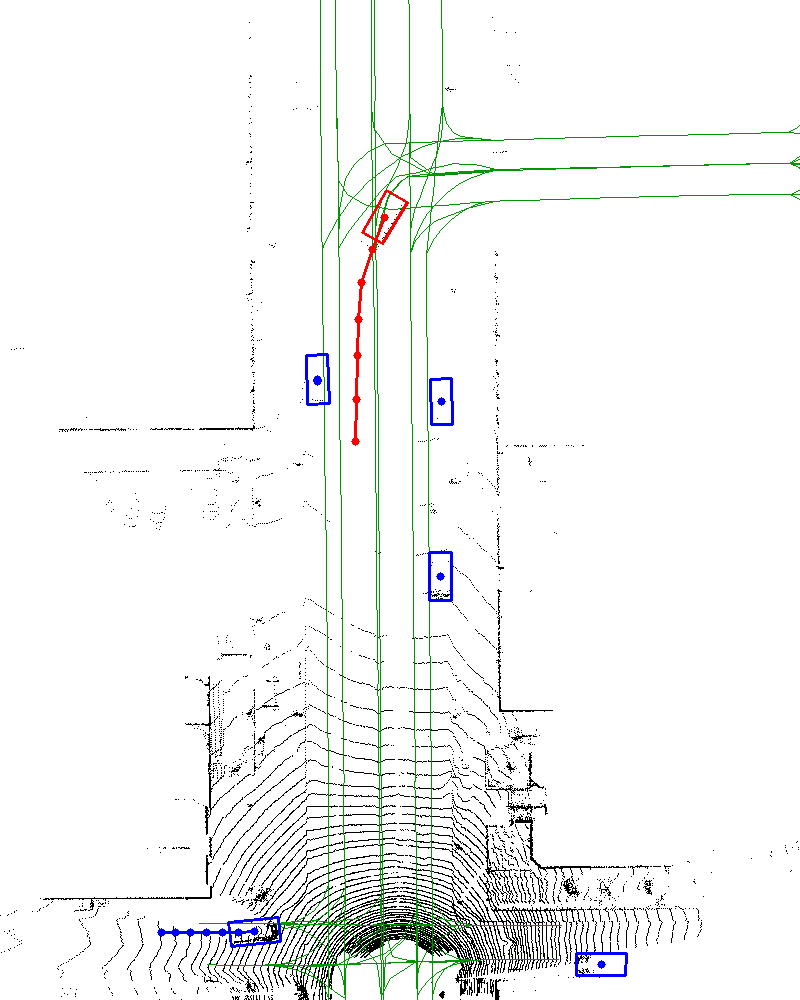}
\end{center}
\vspace{-2mm}
\caption{\bd{Qualitative results of the baseline model without interaction (1st row) and the proposed model (2nd row).} We highlight actors with interactions in \textcolor{red}{red}. With the proposed method, the highlighted vehicles are able to react to other actors and adjust their future trajectory to avoid collision. }
\vspace{-3mm}
\label{fig:visualization}
\end{figure*}

\subsubsection{Comparison with State-of-the-Art}
Table \ref{tab:result} and \ref{tab:result_nuscenes} show the comparison with the state-of-the-art on object detection, trajectory prediction and interaction modeling. Since none of the compared methods uses any form of collision loss, for fair  comparison, we show results of our method both with and without collision loss.
We first compare with previous methods that solve the same end-to-end joint perception and prediction task as ours, \ie, FAF \cite{dpt} and NeuralMP \cite{neuralmp}, on the ATG4D dataset. 
As shown in Table \ref{tab:result}, our model outperforms both approaches by a large margin in all metrics. Specifically, compared to the previous best joint detection and prediction model NeuralMP\cite{neuralmp}, we achieve 4.8\%/6.0\% AP gain at 0.5/0.7 IoU, and 17.9\%/19.3\% relative reduction in ADE/FDE at 90\% recall. In terms of the interaction specific metric TCR, we show that by introducing the Interaction Transformer, we achieve 80.8\% relative reduction at 90\% recall. Adding collision loss further reduces TCR by 56.3\% on a relative basis.
We then compare to SpAGNN \cite{spagnn}, which is the state-of-the-art end-to-end motion forecasting model on nuScenes. SpAGNN takes LiDAR and map as input, and models interaction with graph neural networks, while our model only uses LiDAR as input for nuScenes.
For fair comparisons, in Table \ref{tab:result_nuscenes}, we follow the same metrics as SpAGNN, and demonstrate 34.1\%/22.3\%/74.3\% relative gains on L2@1s, L2@3s, and TCR with IoU 0.1, even without using collision loss.

We also compare with other state-of-the-art interaction modeling approaches, SocialPooling \cite{alahi2016social} and its convolutional variant \cite{convsocial} on both ATG4D and nuScenes, by replacing our Interaction Transformer with these modules. Note that we exploit our powerful multi-sensor fusion backbone on all these baselines. 
As shown in Table \ref{tab:result} and \ref{tab:result_nuscenes}, on ATG4D, even without collision loss, our method is 3.8\%/3.7\%/59.4\% better compared to ConvSocialPool in ADE/FDE/TCR at 90\% detection recall. On nuScenes, at 80\% detection recall, our method is 4.2\%/5.0\%/62.7\% better on L2@1s, L2@3s and TCR with IoU 0.1.
For detection, in addition to the IoU-AP metrics, we also evaluate with the official nuScenes metrics. Our detection model achieves an mAP of 81.4\% on car detection, which is on par with the state-of-the-art. 

\subsubsection{Ablation Studies}
To further analyze the contribution of each module proposed in the paper, we conduct several ablation studies on both the ATG4D-interact testset and the full nuScenes dataset. 
We first provide an ablation study on the high-level model architecture. 
Towards this goal, our baseline model has the same single-stage structure as previous end-to-end models \cite{dpt, neuralmp}, with the only difference being the multi-sensor backbone network. 
We then add our two-stage architecture, the Interaction Transformer, recurrent prediction, and additional refinement on top of the baseline model sequentially. 
As shown in Table \ref{tab:ablation1}, using a two-stage architecture with more specialized feature representation and loss computation for each task brings over 1\% gain in detection AP. Adding the Interaction Transformer improves the detection performance slightly, but significantly improves all prediction metrics on both datasets. The recurrent architecture further pushes the prediction performance; this gain is largely from updated relative location embedding in each prediction time step.
Finally, we add the additional refinement step to recover our full model (without collision loss). This allows the Interaction Transformer to be aware of the spatial relations at the current prediction time step. Compared to the baseline with the same backbone network, our model achieves 6.8\%/7.6\%/83.1\% relative reduction on ADE/FDE/TCR on the ATG4D-interact testset. On nuScenes, we also achieved 4.9\%/7.0\%/54.6\% relative gain on ADE/FDE/TCR. 

Next, we analyze the collision loss. The 2nd and 3rd rows of Table \ref{tab:ablation1} show results of using collision loss without the Interaction Transformer. Collision loss reduces TCR slightly, but at the same time also harms ADE and FDE. In particular, ADE and FDE on ATG4D-interact are 2.0\% and 1.3\% worse compared to the baseline without collision loss. On the other hand, as shown in the last two rows of Table \ref{tab:ablation1}, adding collision loss to our recurrent Interaction Transformer achieves much better results. TCR is further reduced by 55.7\% and 86.6\% on the two datasets, while ADE and FDE are almost unaffected. This indicates that collision loss works better if the network understands spacial relations between actors, as in our Interaction Transformer.

Our last ablation focuses on the design of the Interaction Transformer. For a fair comparison, collision loss is not used here because it harms the baseline model. Our baseline model follows the original Transformer's architecture, which uses a softmax function to normalize the pairwise attention over all other actors. We first add a relative location feature branch to the attention computation, and then improve it by reasoning spatially in each actor's own coordinate system at each prediction time step, instead of the global coordinate system. Finally, we replace the softmax function from the original Transformer with the sigmoid function to remove the constraint that attention scores on all actors must sum to 1. As shown in Table \ref{tab:ablation2}, each modification provides significant gains on all prediction metrics on both datasets. 

\subsubsection{Qualitative Results}
Fig.\ \ref{fig:visualization} shows qualitative results of five driving scenarios, comparing the proposed model and the one-stage baseline without the interaction module (baseline in Table \ref{tab:ablation1}). For fair comparison, neither method uses collision loss. We visualize both detection bounding boxes and estimated future trajectories, and highlight the actors undergoing interactions in red.
With our approach, the target vehicle in the first four scenarios successfully senses the existence of the front vehicle and adjusts its future motion accordingly to avoid collisions.
In contrast, for the baseline, the future prediction of the target vehicle is similar to a simple extrapolation of its past states. As a result, this leads to a collision with the front vehicle. 
In the last scenario, we show that the target vehicle reacts to the parked vehicle and makes a sharper turn to avoid collision.
\section{Conclusion}

We have proposed a novel approach to joint detection and future motion forecasting that takes into account the interactions among the actors. We validate our approach on the challenging ATG4D \cite{dpt} and nuScene \cite{nuscenes2019} datasets and show very significant improvements over the state-of-the-art. 
In the future, we plan to investigate interactions involving other types of actors such as bicyclists and pedestrians, as well as an extension of our approach to jointly perform motion planning in a similar spirit to NeuralMP \cite{neuralmp}.

{\small
\bibliographystyle{IEEEtran}
\bibliography{egbib}
}

\end{document}